# PieAPP: Perceptual Image-Error Assessment through Pairwise Preference

Ekta Prashnani*   Hong Cai*   Yasamin Mostofi   Pradeep Sen

University of California, Santa Barbara

{ekta,hcai,ymostofi,psen}@ece.ucsb.edu

## Abstract

*The ability to estimate the perceptual error between images is an important problem in computer vision with many applications. Although it has been studied extensively, however, no method currently exists that can robustly predict visual differences like humans. Some previous approaches used hand-coded models, but they fail to model the complexity of the human visual system. Others used machine learning to train models on human-labeled datasets, but creating large, high-quality datasets is difficult because people are unable to assign consistent error labels to distorted images. In this paper, we present a new learning-based method that is the first to predict perceptual image error like human observers. Since it is much easier for people to compare two given images and identify the one more similar to a reference than to assign quality scores to each, we propose a new, large-scale dataset labeled with the **probability** that humans will prefer one image over another. We then train a deep-learning model using a novel, **pairwise-learning framework** to predict the preference of one distorted image over the other. Our key observation is that our trained network can then be used separately with only one distorted image and a reference to predict its perceptual error, without ever being trained on explicit human perceptual-error labels. The perceptual error estimated by our new metric, PieAPP, is well-correlated with human opinion. Furthermore, it significantly outperforms existing algorithms, beating the state-of-the-art by almost 3× on our test set in terms of binary error rate, while also generalizing to new kinds of distortions, unlike previous learning-based methods.*

## 1  Introduction

One of the major goals of computer vision is to enable computers to "see" like humans. To this end, a key problem is the automatic computation of the perceptual error (or "distance") of a distorted image with respect to a corresponding reference in a way that is consistent with human observers. A successful solution to this problem would have many applications, including image compression/coding, restoration, and adaptive reconstruction.

*Joint first authors.

This project was supported in part by NSF grants IIS-1321168 and IIS-1619376, as well as a Fall 2017 AI Grant (awarded to Ekta Prashnani).

Because of its importance, this problem, also known as *full-reference image-quality assessment* (FR-IQA) [58], has received significant research attention in the past few decades [5, 11, 23, 29, 30, 32]. The naïve approaches do this by simply computing mathematical distances between the images based on norms such as $\mathcal{L}_2$ or $\mathcal{L}_1$, but these are well-known to be perceptually inaccurate [52]. Others have proposed metrics that try to exploit known aspects of the human visual system (HVS) such as contrast sensitivity [22], high-level structural acuity [52], and masking [48, 51], or use other statistics/features [3,4,14,15,38,44–46,53,54,57]. However, such hand-coded models are fundamentally limited by the difficulty of accurately modeling the complexity of the HVS and therefore do not work well in practice.

To address these limitations, some have proposed IQA methods based on *machine learning* to learn more sophisticated models [19]. Although many learning-based methods use hand-crafted image features [8, 10, 18, 20, 21, 31, 33, 35, 40], recent methods (including ours) apply *deep-learning* to FR-IQA to learn features automatically [7, 17, 25]. However, the accuracy of all existing learning-based methods depends on the size and quality of the datasets they are trained on, and existing IQA datasets are small and noisy. For instance, many datasets [16, 26, 27, 34, 47, 49, 55] are labeled using a *mean opinion score* (MOS) where each user gives the distorted image a subjective quality rating (e.g., 0 = "bad", 10 = "excellent"). These individual scores are then averaged in an attempt to reduce noise. Unfortunately, creating a good IQA dataset in this fashion is difficult because humans cannot assign quality or error labels to a distorted image consistently, even when comparing to a reference (e.g., try rating the images in Fig.1 from 0 to 10!).

Other datasets (e.g., TID2008 [43] and TID2013 [42]) leverage the fact that it is much easier for people to select which image from a distorted pair is closer to the reference than to assign them quality scores. To translate user preferences into quality scores, they then take a set of distorted images and use a *Swiss tournament* [1] to assign scores to each. However, this approach has the fundamental problem that the *same* distorted image could have varying scores in different sets (see supplementary for examples in TID2008 and TID2013). Moreover, the number of images and distortion types in all of these datasets is very limited. The largest dataset we know of (TID2013) has only 25 images and 24 distortions, which hardly qualifies as "big-data" for machine learning. Thus, methods trained on these datasets have limited generalizability to new distortions, as we will show later. Because of these limitations, no method currently exists that can predict perceptual error like human observers, even for easy examples such as the one in Fig. 1. Here, although the answer is obvious to most people, all existing FR-IQA methods give the wrong answer, confirming that this problem is clearly far from solved.

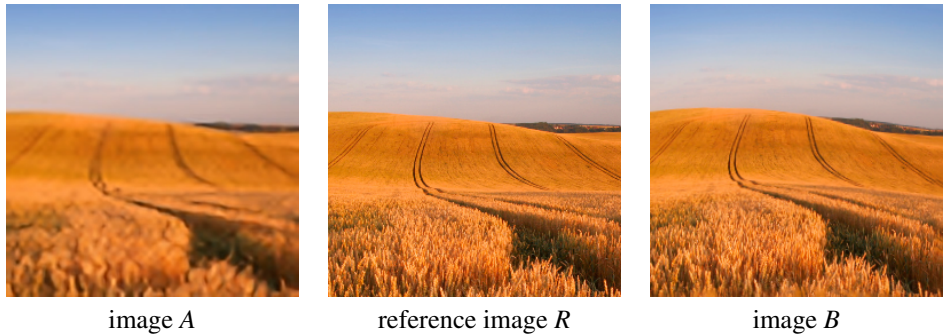

| METHOD | A | B |
|---|---|---|
| **Humans** | 0.120 | **0.880** |
| MAE | **8.764** | 16.723 |
| RMSE | **15.511** | 28.941 |
| SSIM | **0.692** | 0.502 |
| MS-SSIM | **0.940** | 0.707 |
| PSNR-HMA | **27.795** | 19.763 |
| FSIMc | **0.757** | 0.720 |
| SFF | **0.952** | 0.803 |
| GMSD | **0.101** | 0.176 |
| VSI | **0.952** | 0.941 |
| SCQI | **0.996** | 0.996 |
| Lukin et al. [35] | **3.213** | 1.504 |
| Bosse et al. [7] | **31.360** | 36.192 |
| Kim et al. [25] | **0.414** | 0.283 |
| **PieAPP (error)** | 2.541 | **0.520** |
| **PieAPP (prob.)** | 0.117 | **0.883** |

Figure 1: Which image, *A* or *B*, is more similar to the reference *R*? This is an example of a pairwise image comparison where most people have no difficulty determining which image is closer. In this case, according to our Amazon Mechanical Turk (MTurk) experiments, 88% of people prefer image *B*. Despite this simple visual task, 13 image quality assessment (IQA) methods–including both popular and state-of-the-art approaches–fail to predict the image that is visually closer to the reference. On the other hand, our proposed PieAPP error metric correctly predicts that *B* is better with a preference probability of 88.3% (or equivalently, an error of 2.541 for *A* and 0.520 for *B*, with the reference having an error of 0). Note that neither the reference image nor the distortion types present were available in our training set.

In this paper, we make critical strides towards solving this problem by proposing a novel framework for learning perceptual image error as well as a new, corresponding dataset that is larger and of higher quality than previous ones. We first describe the dataset, since it motivates our framework. Rather than asking people to label images with a subjective quality score, we exploit the fact that it is much easier for humans to select which of two images is closer to a reference. However, unlike the TID datasets [42, 43], we do not explicitly convert this preference into a quality score, since approaches such as Swiss tournaments introduce errors and do not scale. Instead, we simply label the pairs by the percentage of people who preferred image *A* over *B* (e.g., a value of 50% indicates that both images are equally "distant" from the reference). By using this pairwise probability of preference as ground-truth labels, our dataset can be larger and more robust than previous IQA datasets.

Next, our proposed *pairwise-learning framework* trains an error-estimation function using the probability labels in our dataset. To do this, we input the distorted images (*A*,*B*) and the corresponding reference, *R*, into a pair of identical error-estimation functions which output the perceptual-error scores for *A* and *B*. The choice for the error-estimation function is flexible, and in this paper we propose a new deep convolutional neural network (DCNN) for it. The errors of *A* and *B* are then used to compute the predicted probability of preference for the image pair. Once our system, which we call *PieAPP*, is trained using the pairwise probabilities, we can use the learned error-estimation function on a single image *A* and a reference *R* to compute the perceptual error of *A* with respect to *R*. This trick allows us to quantify the perceived error of a distorted image with respect to a reference, even though **our system was never explicitly trained with hand-labeled, perceptual-error scores.**

The combination of our novel, pairwise-learning framework and new dataset results in a significant improvement in perceptual image-error assessment, and can also be used to further improve existing learning-based IQA methods. Interested readers can find our code and trained models at https://doi.org/10.7919/F4GX48M7.

## 2 Pairwise learning of perceptual image error

Existing IQA datasets (e.g., LIVE [47], TID2008 [43], CSIQ [26], and TID2013 [42]) suffer from either the unreliable human rating of image quality or the set-dependence of Swiss tournaments. Unlike these previous datasets, our proposed dataset focuses exclusively on the probability of pairwise preference. In other words, given two distorted versions (*A* and *B*) of reference image *R*, subjects are asked to select the one that looks more similar to *R*. We then store the percentage of people who selected image *A* over *B* as the ground-truth label for this pair, which we call the *probability of preference* of *A* over *B* (written as $p_{AB}$). This approach is more robust because it is easier to identify the closer image than to assign quality scores, and does not suffer from set-dependency or scalability issues like Swiss tournaments since we never label the images with quality scores.

The challenge is how to use these probabilistic preference labels to estimate the perceptual-error scores of individual images compared to the reference. To do this, we assume as shown in Fig. 2 that all distorted versions of a reference image can be mapped to a 1-D "perceptual-error" axis (as is common in IQA), with the reference at the origin and distorted versions placed at varying distances from the origin based on their perceptual error (images that are more perceptually similar to the reference are closer, others farther away). Note that since each reference image has its own quality axis, comparing a distorted version of one reference to that of another does not make logical sense.

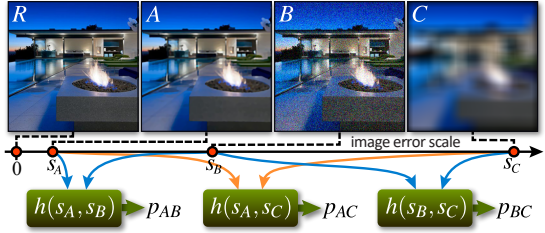

Figure 2: Like all IQA methods, we assume distorted images can be placed on a linear scale based on their underlying perceptual-error scores (e.g., $s_A, s_B, s_C$) with respect to the reference. In our case, we map the reference to have 0 error. We assume the probability of preferring distorted image A over B can be computed by applying a function $h$ to their errors, e.g., $p_{AB} = h(s_A, s_B)$.

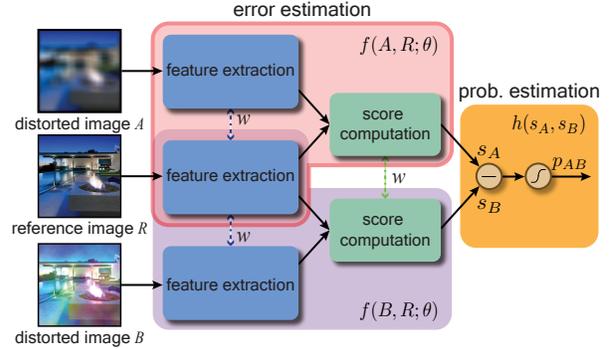

Figure 3: Our pairwise-learning framework consists of error- and probability-estimation blocks. In our implementation, the error-estimation function $f$ has two weight-shared feature-extraction (FE) networks that take in reference $R$ and a distorted input ($A$ or $B$), and a score-computation (SC) network that uses the extracted features from each image to compute the perceptual-error score (see Fig. 4 for more details). Note that the FE block for $R$ is shared between $f(A, R; \theta)$ and $f(B, R; \theta)$. The computed perceptual-error scores for $A$ and $B$ ($s_A$ and $s_B$) are then passed to the probability-estimation function $h$, which implements the Bradley-Terry (BT) model (Eq. 1) and outputs the probability of preferring $A$ over $B$.

Given this axis, we assume there is a function $h$ which takes the perceptual-error scores of $A$ and $B$ (denoted by $s_A$ and $s_B$, respectively), and computes the probability of preferring $A$ over $B$: $p_{AB} = h(s_A, s_B)$. In this paper, we use the Bradley-Terry (BT) sigmoid model [9] for $h$, since it has successfully modeled human responses for pairwise comparisons in other applications [13, 39]:[1]

$$p_{AB} = h(s_A, s_B) = \frac{1}{1 + e^{s_A - s_B}}. \quad (1)$$

Unlike the standard BT model, the exponent here is negated so that lower scores are assigned to images visually closer to the reference. Given this, our goal is then to learn a function $f$ that maps a distorted image to its perceptual error with respect to the reference, constrained by the observed probabilities of preference. More specifically, we propose a general optimization framework to train $f$ as follows:

$$\hat{\theta} = \underset{\theta}{\text{argmin}} \frac{1}{T} \sum_{i=1}^{T} \| h(f(A_i, R_i; \theta), f(B_i, R_i; \theta)) - p_{AB,i} \|_2^2, \quad (2)$$

where $\theta$ denotes the parameters of the image error-estimation function $f$, $p_{AB,i}$ is the ground-truth probability of preference based on human responses, and $T$ is the total number of training pairs. If the training data is fitted correctly and is sufficient in terms of images and distortions, Eq. 2 will train $f$ to estimate the underlying perceptual-error scores for every image so that their relative spacing on the image quality scale will match their pairwise probabilities (enforced by Eq. 1), with images that are closer to the reference having smaller numbers. These underlying perceptual errors are estimated up to an additive constant, as only the *relative distances* between images are constrained by Eq. 1. We discuss how to account for this constant by setting the error of the reference with itself to 0 in Sec. 3.

---
[1] We empirically verify that BT is consistent with our collected human responses in Sec. 5.1.

To learn the error-estimation function $f$, we propose a novel *pairwise-learning framework*, shown in Fig. 3. The inputs to our system are sets of three images ($A$, $B$, and $R$), and the output is the probability of preferring $A$ over $B$ with respect to $R$. Our framework has two main learning blocks, $f(A, R; \theta)$ and $f(B, R; \theta)$, that compute the perceptual error of each image. The estimated errors $s_A$ and $s_B$ are then subtracted and fed through a sigmoid that implements the BT model in Eq. 1 (function $h$) to predict the probability of preferring $A$ over $B$. The entire system can then be trained by backpropagating the squared $\mathcal{L}_2$ error between the predicted probabilities and the ground-truth human preference labels to minimize Eq. 2.

At this point, we simply need to make sure we have an expressive computational model for $f$ as well as a large dataset with a rich variety of images and distortion types. To model $f$, we propose a new DCNN-based architecture which we describe in Sec. 3. For the dataset, we propose a new large-scale image distortion dataset with probabilistic pairwise human comparison labels as discussed in Sec. 4.

## 3  New DCNN for image-error assessment

Before describing our implementation for function $f$, we note that our pairwise-learning framework is general and can by used to train *any* learning model for error computation by simply replacing $f(A, R; \theta)$ and $f(B, R; \theta)$. In fact, we show in Sec. 5.4 how the performance of Bosse et al. [7]'s and Kim et al. [25]'s architectures is considerably improved when integrated into our framework. Furthermore, once trained on our framework, the **error-estimation function $f$ can be used by itself to compute the perceptual error of individual images with respect to a reference**. Indeed, this is how we get all the results in the paper.

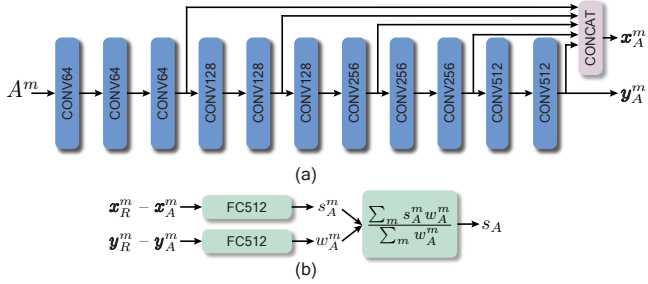

(a)

(b)

Figure 4: Our DCNN implementation of the error-estimation function $f$. (a) The feature-extraction (FE) subnet of $f$ has 11 convolutional (CONV) layers with skip connections to compute the features for an input patch $A^m$. The number after "CONV" indicates the number of feature maps. Each layer has $3\times 3$ filters and a non-linear ReLU, with $2\times 2$ max-pooling after every even layer. (b) The score-computation (SC) subnet uses two fully-connected (FC) networks (each with 1 hidden layer with 512 neurons) to compute patch-wise weights and errors, followed by weighted averaging over all patches to compute the final image score.

In our implementation, the error-estimation block $f$ consists of two kinds of subnetworks (subnets, for short). There are three identical, weight-shared feature-extraction (FE) subnets (one for each input image), and two weight-shared score-computation (SC) subnets that compute the perceptual-error scores for $A$ and $B$. Together, two FE and one SC subnets comprise the error-estimation function $f$. As is common in other IQA algorithms [7, 56, 57], we compute errors on a patchwise basis by feeding corresponding patches from $A$, $B$, and $R$ through the FE and SC subnets, and aggregate them to obtain the overall errors, $s_A$ and $s_B$.

Figure 4 shows the details of our implementation of function $f$. The three (weight-shared) FE subnets each consist of 11 convolutional (CONV) layers (Fig. 4a). For each set of input patches ($A^m$, $B^m$, and $R^m$, where $m$ is the patch index), the corresponding feature maps from the FE CONV layers at different depths are flattened and concatenated into feature vectors $x_A^m$, $x_B^m$, and $x_R^m$. Using features from multiple layers has two advantages: 1) multiple CONV layers contain features from different scales of the input image, thereby leveraging both high-level and low-level features for error score computation, and 2) skip connections enable better gradient backpropagation through the network.

Once these feature vectors are computed by the FE subnet, the *differences* between the corresponding feature vectors of the distorted and reference patches are fed into the SC subnet (Fig. 4b). Each SC subnet consists of two fully-connected (FC) networks. The first FC network takes in the multi-layer feature difference (i.e., $x_R^m - x_A^m$), and predicts the patchwise error ($s_A^m$). These are aggregated using weighted averaging to compute the overall image error ($s_A$), where the weight for each patch $w_A^m$ is computed using the second FC network (similar to Bosse et al. [7]). This network uses the feature difference from the last CONV layer of the FE subnet as input (denoted as $y_R^m - y_A^m$ for $A$ and $R$), since the weight for a patch is akin to the higher-level patch saliency [7, 56, 57] captured by deeper CONV layers.

Feeding the feature differences to the SC subnet ensures that when estimating the perceptual error for a reference image (i.e., $A = R$), the SC block would receive $x_R^m - x_A^m = \mathbf{0}$ as input. The system would therefore output a constant value which is invariant to the reference image, caused by the bias terms in the fully-connected networks in the SC subnet. By subtracting this constant from the predicted error, we ensure that the "origin" of the quality axis is always positioned at 0 for each reference image.

To train our proposed architecture, we adopt a random patch-sampling strategy [7], which prevents over-fitting and improves learning. At every training iteration, we randomly sample 36 patches of size $64\times 64$ from our training images which are of size $256\times 256$. The density of our patch sampling ensures that any pixel in the input image is included in at least one patch with a high probability (0.900). This is in contrast with earlier approaches [7], where patches are sampled sparsely and there is only a 0.154 probability that a specific pixel in an image will be in one of the sampled patches. This makes it harder to learn a good perceptual-error metric.[2] At test time, we randomly sample $1{,}024$ patches for each image to compute the perceptual error. We now describe our dataset for training the proposed framework.

## 4 Large-scale image distortion dataset

As discussed earlier, existing IQA datasets [26, 42, 43, 47] suffer from many problems, such as unreliable quality labels and a limited variety of image contents and distortions. For example, they do not contain many important distortions that appear in real-world computer vision and image processing applications, such as artifacts from deblurring or dehazing. As a result, training high-quality perceptual-error metrics with these datasets is difficult, if not impossible.

To address these problems (and train our proposed system), we have created our own large-scale dataset, labeled with pairwise probability of preference, that includes a wide variety of image distortions. Furthermore, we also built a test set with a large number of images and distortion types that do not overlap with the training set, allowing a rigorous evaluation of the generalizability of IQA algorithms.

Table 1 compares our proposed dataset with the four largest existing IQA datasets.[3] Our dataset is substantially bigger than all these existing IQA datasets *combined* in terms of the number of reference images, the number of distortion types, and the total number of distorted images. We next discuss the composition of our dataset.

---

[2]See supplementary file for a detailed analysis.
[3]See Chandler et al. [11] for a complete list of existing IQA datasets.

| Dataset | Ref. images | Distortions | Distorted images |
|---|---|---|---|
| LIVE [47] | 29 | 5 | 779 |
| CSIQ [26] | 30 | 6 | 866 |
| TID2008 [43] | 25 | 17 | 1,700 |
| TID2013 [42] | 25 | 24 | 3,000 |
| **Our dataset** | **200** | **75** | **20,280** |

Table 1: Comparison of the four largest IQA dataset and our proposed dataset, in terms of the number of reference images, the number of distortions, and the number of distorted images.

**Reference images:** The proposed dataset contains 200 unique reference images (160 reference images are used for training and 40 for testing), which are selected from the Waterloo Exploration Database [36, 37] because of its high-quality images. The selected reference images are representative of a wide variety of real-world content. Currently, the image size in our dataset is $256 \times 256$, which is a popular size in computer vision and image processing applications. This size also enables crowdsourced workers to evaluate the images without scrolling the screen. However, we note that since our architecture samples patches from the images, it can work on input images of various sizes.

**Image distortions:** In our proposed dataset, we have included a total of 75 distortions, with a total of 44 distortions in the training set, and 31 in the test set which are distinct from the training set.[4] More specifically, our set of image distortions spans the following categories: **1)** common image artifacts (e.g., additive Gaussian noise, speckle noise); **2)** distortions that capture important aspects of the HVS (e.g., non-eccentricity, contrast sensitivity); and **3)** complex artifacts from computer vision and image processing algorithms (e.g., deblurring, denoising, super-resolution, compression, geometric transformations, color transformations, and reconstruction). Although recent IQA datasets cover some of the distortions in categories 1 and 2, they do not contain many distortions from category 3 even though they are important to computer vision and image processing. We refer the readers to the supplementary file for a complete list of the training and test distortions in our dataset.

### 4.1 Training set

We select 160 reference images and 44 distortions for training PieAPP. Each training example is a pairwise comparison consisting of a reference image $R$, two distorted versions $A$ and $B$, along with a label $\tilde{p}_{AB}$, which is the estimated probabilistic human preference based on collected human data (see Sec. 4.3). For each reference image $R$, we design two kinds of pairwise comparisons: *inter-type* and *intra-type*. In an inter-type comparison, $A$ and $B$ are generated by applying two different types of distortions to $R$. For each reference image, there are 4 groups of inter-type comparisons, each containing 15 distorted images generated using 15 randomly-sampled distortions.[5] On the other hand, in an intra-type comparison, $A$ and $B$ are generated by applying the same distortion to $R$ with different parameters. For each reference image, there are 21 groups of intra-type comparisons, containing 3 distorted images generated using the same distortion with different parameter settings. The exhaustive pairwise comparisons within each group (both inter-type and intra-type) and the corresponding human labels $\tilde{p}_{AB}$ are then used as the training data. Overall, there are a total of 77,280 pairwise comparisons for training (67,200 inter-type and 10,080 intra-type). Inter-distortion comparisons allow us to capture human preference across different distortion types and are more challenging than the intra-distortion comparisons due to a larger variety of pairwise combinations and the difficulty in comparing images with different distortion types. We therefore devote a larger proportion of our dataset to inter-distortion comparisons.

[4]In contrast, most previous learning-based IQA algorithms test on the same distortions that they train on. Even in the "cross-dataset" tests presented in previous papers, there is a significant overlap between the training and test distortions. This makes it impossible to tell whether previous learning-based algorithms would work for new, unseen distortions.

### 4.2 Test set of unseen distortions and images

The test set contains 40 reference images and 31 distortions, which are representative of a variety of image contents and visual effects. None of these images and distortions are in the training set. For each reference image, there are 15 distorted images with randomly-sampled distortions (sampled to ensure that the test set has both inter and intra-type comparisons). Probabilistic labels are assigned to the exhaustive pairwise comparisons of the 15 distorted images for each reference. The test set then contains a total of 4,200 distorted image pairs (105 per reference image).

### 4.3 Amazon Mechanical Turk data collection

We use Amazon Mechanical Turk (MTurk) to collect human responses for both the training and test pairs. In each pairwise image comparison, the MTurk user is presented with distorted images ($A$ and $B$) and the reference $R$. The user is asked to select the image that he/she considers more similar to the reference.[6] However, we need to collect a sufficient number of responses per pair to accurately estimate $p_{AB}$. Furthermore, we need to do this for $77,280$ training pairs and $4,200$ test pairs, resulting in a large number of MTurk inquiries and a prohibitive cost. In the next two sections, we show how to avoid this problem by analyzing the number of responses needed per image pair to statistically estimate its $p_{AB}$ accurately, and then showing how to use a maximum likelihood (ML) estimator to accurately label a larger set of pairs based on a smaller set of acquired labels.

[5]The choice of 15 is based on a balance between properly sampling the training distortions and the cost of obtaining labels in an inter-type group.
[6]Details on MTurk experiments and interface are in the supplementary.

### 4.3.1 Number of responses per comparison

We model the human response as a Bernoulli random variable $\nu$ with a success probability $p_{AB}$, which is the probability of a person preferring $A$ over $B$. Given $n$ human responses $\nu_i, i=1,...,n$, we can estimate $p_{AB}$ by $\tilde{p}_{AB} = \frac{1}{n}\sum_{i=1}^{n}\nu_i$. We must choose $n$ such that $\text{prob}(|\tilde{p}_{AB} - p_{AB}| \leq \eta) \geq P_{\text{target}}$, for a target $P_{\text{target}}$ and tolerance $\eta$. It can be easily confirmed (see supplementary) that by choosing $n = 40$ and $\eta = 0.15$, we can achieve a reasonable $P_{\text{target}} \geq 0.94$. Therefore, we collect 40 responses for each pairwise comparison in the training and test sets.

### 4.3.2 Statistical estimation of human preference

Collecting 40 MTurk responses for 77,280 pairs of training images is expensive. Thus, we use statistical modeling to estimate the missing human labels based on a subset of the exhaustive pairwise comparison data [50]. To see how, suppose we need to estimate $p_{AB}$ for all possible pairs of $N$ images (e.g., $N = 15$ in each inter-type group) with perceptual-error scores $\boldsymbol{s} = [s_1,...,s_N]$. We denote the human responses by a count matrix $C = \{c_{i,j}\}$, where $c_{i,j}$ is the number of times image $i$ is preferred over image $j$. The scores can then be obtained by solving an ML estimation problem [50]: $\boldsymbol{s}^\star = \arg\max \sum_{i,j} c_{i,j} \log S(s_i - s_j)$, where $S(.)$ is the sigmoid function of the BT model for human responses (see Sec. 2). We solve this optimization problem via gradient descent.

However, we must query a sufficient number of pairwise comparisons so that the optimal solution recovers the underlying true scores. It is sufficient to query a subset of all the possible comparisons as long as each image appears in at least $k$ comparisons ($k < N-1$) presented to the humans, where $k$ can be determined empirically [39, 59]. Our empirical analysis reveals that $k = 10$ is sufficient as the binary error rate over the estimated part of the subset becomes 0.0006. More details are included in the supplementary file.

ML estimation reduces the number of pairs that need labeling considerably, from 81,480 to 62,280 (23.56% reduction). Note that we only do this for the training set and not for the test set (we query 40 human responses for each pairwise comparison in the test set), which ensures that no test errors are caused by possible ML estimation errors.

## 5 Results

We implemented the system presented in Sec. 3 in TensorFlow [2], and trained it on the training set described in Sec. 4 for 300K iterations (2 days) on an NVIDIA Titan X GPU. In this section, we evaluate the performance of PieAPP and compare it with popular and state-of-the-art IQA methods on both our proposed test set (Sec. 4.2) and two existing datasets (CSIQ [26] and TID2013 [42]). Since there are many recent learning-based algorithms (e.g., [17, 24, 28]), in this paper we only compare against the three that performed the best on established datasets based on their published results [7, 25]. We show comparisons against other methods and on other datasets in the supplementary.

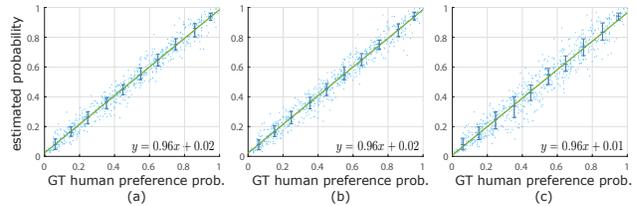

Figure 5: Ground-truth (GT) human-preference probabilities vs. estimated probabilities. **(a)** Estimating probabilities based on *all* pairwise comparisons with 100 human responses per pair to validate the BT model. **(b)** Estimating probabilities with only 10 comparisons per image (still 100 human responses each). **(c)** Estimating probabilities based on only 10 comparisons per image with only 40 responses per pair. The last two plots show the accuracy of using ML estimation to fill in the missing labels (see Sec. 4.3.2). The blue segments near the fitted line indicate the 25$^{\text{th}}$ and 75$^{\text{th}}$ percentile of the estimated probabilities within each 0.1 bin of GT probabilities. The fitted line is close to $y = x$ as shown by the equations on the bottom-right.

We begin by validating the BT model for our data and showing that ML estimation can accurately fill in missing human labels (Sec. 5.1). Next, we compare the performance of PieAPP on our test set against that of popular and state-of-the-art IQA methods (Sec. 5.2), with our main results presented in Table 2. We also test our learning model (given by the error-estimation block $f$) on other datasets by training it on them (Sec. 5.3, Table 3). Finally, we show that existing learning-based methods can be improved using our pairwise-learning framework (Sec. 5.4, Table. 4).

### 5.1 Consistency of the BT model and the estimation of probabilistic human preference

We begin by showing the validity of our assumptions that: 1) the Bradley-Terry (BT) model accurately accounts for human decision-making for pairwise image comparisons, and 2) ML estimation can be used to accurately fill in the human preference probability for pairs without human labels. To do these experiments, we first collect exhaustive pairwise labels for 6 sets of 15 distorted images (total 630 image pairs), with 100 human responses for each comparison, which ensures that the probability labels are highly reliable ($P_{\text{target}} = 0.972$ and $\eta = 0.11$; see Sec. 4.3.1).

To validate the BT model, we estimate the scores for all images in a set with ML estimation, using the ground-truth labels of all the pairs.[7] Using the estimated scores, we compute the preference probability for each pair based on BT (Eq. 1), which effectively tests whether the BT scores can "fit" the measured pairwise probabilities. Indeed, Fig. 5a shows that the relationship between the ground-truth and the estimated probabilities is close to identity, which indicates that BT is a good fit for human decision-making.

We then validate the ML estimation when not all pair-wise comparisons are labeled. We estimate the scores using 10 ground-truth comparisons per image in each set ($k = 10$; see Sec. 4.3.2), instead of using all the pairwise comparisons like before. Fig. 5b shows that we have a close-to-identity relationship with a negligible increase of noise, indicating the good accuracy of the ML-estimation process.

---

[7]The is the same estimation as in Sec 4.3.2, but with the ground-truth labels for all the pairs in each set.

| METHOD | KRCC | | PLCC | SRCC |
|---|---|---|---|---|
| | $\tilde{p}_{AB} \in [0,1]$ | $\tilde{p}_{AB} \notin [0.35, 0.65]$ | | |
| MAE | 0.252 | 0.289 | 0.302 | 0.302 |
| RMSE | 0.289 | 0.339 | 0.324 | 0.351 |
| SSIM | 0.272 | 0.323 | 0.245 | 0.316 |
| MS-SSIM | 0.275 | 0.325 | 0.051 | 0.321 |
| GMSD | 0.250 | 0.291 | 0.242 | 0.297 |
| VSI | 0.337 | 0.395 | 0.344 | 0.393 |
| PSNR-HMA | 0.245 | 0.274 | 0.310 | 0.281 |
| FSIMc | 0.322 | 0.377 | 0.481 | 0.378 |
| SFF | 0.258 | 0.295 | 0.025 | 0.305 |
| SCQI | 0.303 | 0.364 | 0.267 | 0.360 |
| DOG-SSIMc | 0.263 | 0.320 | 0.417 | 0.464 |
| Lukin et al. | 0.290 | 0.396 | 0.496 | 0.386 |
| Kim et al. | 0.211 | 0.240 | 0.172 | 0.252 |
| Bosse et al. (NR) | 0.269 | 0.353 | 0.439 | 0.352 |
| Bosse et al. (FR) | 0.414 | 0.503 | 0.568 | 0.537 |
| **Our method (PieAPP)** | **0.668** | **0.815** | **0.842** | **0.831** |

Table 2: Performance of our approach compared to existing IQA methods on our test set. PieAPP beats all the state-of-the-art methods because the test set contains many different (and complex) distortions not found in standard IQA datasets.

| METHOD | CSIQ [26] | | | TID2013 [42] | | |
|---|---|---|---|---|---|---|
| | KRCC | PLCC | SRCC | KRCC | PLCC | SRCC |
| MAE | 0.639 | 0.644 | 0.813 | 0.351 | 0.294 | 0.484 |
| RMSE | 0.617 | 0.752 | 0.783 | 0.327 | 0.358 | 0.453 |
| SSIM | 0.691 | 0.861 | 0.876 | 0.464 | 0.691 | 0.637 |
| MS-SSIM | 0.739 | 0.899 | 0.913 | 0.608 | 0.833 | 0.786 |
| GMSD | 0.812 | 0.954 | 0.957 | 0.634 | 0.859 | 0.804 |
| VSI | 0.786 | 0.928 | 0.942 | 0.718 | 0.900 | 0.897 |
| PSNR-HMA | 0.780 | 0.888 | 0.922 | 0.632 | 0.802 | 0.813 |
| FSIMc | 0.769 | 0.919 | 0.931 | 0.667 | 0.877 | 0.851 |
| SFF | 0.828 | 0.964 | 0.963 | 0.658 | 0.871 | 0.851 |
| SCQI | 0.787 | 0.927 | 0.943 | 0.733 | 0.907 | 0.905 |
| DOG-SSIMc | 0.813 | 0.943 | 0.954 | 0.768 | 0.934 | 0.926 |
| Lukin et al. | – | – | – | 0.770 | – | 0.930 |
| Kim et al. | – | 0.965 | 0.961 | – | **0.947** | 0.939 |
| Bosse et al. (NR) | – | – | – | – | 0.787 | 0.761 |
| Bosse et al. (FR) | – | – | – | 0.780 | 0.946 | 0.940 |
| **Error-estimation $f$** | **0.881** | **0.975** | **0.973** | **0.804** | 0.946 | **0.945** |

Table 3: Comparison on two standard IQA datasets (CSIQ [26] and TID2013 [42]). For all the learning methods, we used the numbers directly provided by the authors (dashes "–" indicate numbers were not provided). For a fair comparison, we used only one error-estimation block of our pairwise-learning framework (function $f$) trained directly on the MOS labels of each dataset. The performance of PieAPP on these datasets, when trained with our pairwise-learning framework, is shown in Table 4.

Finally, we reduce the number of human responses per comparison: we again use 10 comparisons per image in each set, but this time with only 40 responses per comparison (as we did for our entire training set), instead of the 100 responses we used previously. Fig. 5c shows that the noise has increased slightly but the fit to the ground-truth labels is still quite good. Hence, this validates the way we supplemented our hand-labeled data using ML estimation.

## 5.2 Performance on our unseen test set

We now compare the performance of our proposed PieAPP metric to that of other IQA methods on our test set, where the images and distortion types are completely disjoint from the training set. This tests the generalizability of the various approaches to new distortions and image content. We compare the methods using the following evaluation criteria:

**1. Accuracy of predicted quality (or perceptual error):** As discussed in Sec. 5.1, we obtain the ground-truth scores through ML estimation, using the ground-truth preference labels for all the pairs in the test set. As is typically done in IQA papers, we compute the *Pearson's linear correlation coefficient* (PLCC) to assess the correlation between the magnitudes of the scores predicted by the IQA method and the ground-truth scores.[8] We also use the *Spearman's rank correlation coefficient* (SRCC) to assess the agreement of ranking of images based on the predicted scores.

**2. Accuracy of predicted pairwise preference:** IQA methods are often used to tell which distorted image, *A* or *B*, is closer to a reference. Therefore, we want to know the *binary error rate* (BER), the percentage of test set pairs predicted incorrectly. We report the *Kendall's rank correlation coefficient* (KRCC), which is related to the BER by KRCC = $1 - 2$BER. Since this is less meaningful when human preference is not strong (i.e., $\tilde{p}_{AB} \in [0.35, 0.65]$), we show numbers for both the full range and $\tilde{p}_{AB} \notin [0.35, 0.65]$.

To compare, we also test **1) model-based methods:** Mean Absolute Error (MAE), Root Mean Squared Error (RMSE), SSIM [52], MS-SSIM [53], GMSD [54], VSI [56], PSNR-HMA [41], FSIMc [57], SFF [12], and SCQI [4], and **2) learning-based methods:** DOG-SSIMc [40], Lukin et al. [35], Kim et al. [25], and Bosse et al. [6, 7].[9] Bosse et al. proposed full-reference (FR) and no-reference (NR) versions of their method and we test against both. In all cases, we use the code/models released by the authors, except for Kim et al. [25], whose trained model is not publicly available. Therefore, we used the source code provided by the authors to train their model as described in their paper, and validated it by getting their reported results.

As Table 2 shows, our proposed method significantly outperforms existing state-of-the-art IQA methods. Our PLCC and SRCC are 0.842 and 0.831, respectively, outperforming the second-best method, Bosse et al. (FR) [7], by 48.24% and 54.75%, respectively. This shows that our predicted perceptual error is considerably more consistent with the ground-truth scores than state-of-the-art methods. Furthermore, the KRCC of our approach over the entire test set ($\tilde{p}_{AB} \in [0,1]$) is 0.668, and is 0.815 when $\tilde{p}_{AB} \notin [0.35, 0.65]$ (i.e., when there is a stronger preference by humans). This is a significant improvement over Bosse et al. (FR) [7] of 61.35% and 62.03%, respectively. Translating these to binary error rate (BER), we see that our method has a BER of 9.25% when $\tilde{p}_{AB} \notin [0.35, 0.65]$, while Bosse et al. has a BER of 24.85% in the same range. This means that the best IQA method to date gets almost 25% of the pairwise comparisons wrong, but our approach offers a 2.7× improvement. The fact that we get these results on our test set (which is disjoint from the training set) indicates that PieAPP is capable of generalizing to new image distortions and content much better than existing methods.

---
[8] For existing IQA methods, the PLCC on our test set is computed after fitting the predicted scores to the ground-truth scores via a nonlinear regression as is commonly done [47].

[9] See supplementary for brief descriptions of these methods, as well as comparisons to other IQA methods.

| Our PLF Modifications: | Our test set | | | | CSIQ [26] | | | TID2013 [42] | | |
|---|---|---|---|---|---|---|---|---|---|---|
| | KRCC | | PLCC | SRCC | KRCC | PLCC | SRCC | KRCC | PLCC | SRCC |
| | $\tilde{p}_{AB} \in [0,1]$ | $\tilde{p}_{AB} \notin [0.35, 0.65]$ | | | | | | | | |
| PLF + Kim et al. | 0.491 | 0.608 | 0.654 | 0.632 | 0.708 | **0.863** | 0.873 | 0.649 | 0.795 | 0.837 |
| PLF + Bosse et al. (NR) | 0.470 | 0.593 | 0.590 | 0.593 | 0.663 | 0.809 | 0.842 | 0.654 | 0.781 | 0.831 |
| PLF + Bosse et al. (FR) | 0.588 | 0.729 | 0.734 | 0.748 | 0.739 | 0.844 | 0.898 | 0.682 | 0.828 | 0.859 |
| **Our method (PieAPP)** | **0.668** | **0.815** | **0.842** | **0.831** | **0.754** | 0.842 | **0.907** | **0.710** | **0.836** | **0.875** |

Table 4: Our novel pairwise-learning framework (PLF) can also be used to improve the quality of existing learning-based IQA methods. Here, we replaced the error-estimating $f$ blocks in our pairwise-learning framework (see Fig. 3) with the learning models of Kim et al. [25] and Bosse et al. [7]. We then trained their architectures using our pairwise-learning process on our training set. As a result, the algorithms improved considerably on our test set, as can be seen by comparing these results to those of their original versions in Table 2. Furthermore, we also evaluated these methods on the standard CSIQ [26] and TID13 [42] datasets using the original MOS labels for ground truth.

### 5.3 Testing our architecture on other IQA datasets

For completeness, we also compare our architecture against other IQA methods on two of the largest existing IQA datasets, CSIQ [26] and TID2013 [42].[10] Since these have MOS labels which are noisy with respect to ground-truth preferences, we trained our error-estimation function from the pairwise-learning framework (i.e., $f(A, R; \theta)$ in Fig. 3) on the MOS labels of CSIQ and TID2013, respectively.[11] This allows us to directly compare our DCNN architecture to existing methods on the datasets they were designed for.

For these experiments, we randomly split the datasets into 60% training, 20% validation, and 20% test set, and report the performance averaged over 5 such random splits, as is usual. The standard deviation of the correlation coefficients on the test sets of these five random splits is at most 0.008 on CSIQ and at most 0.005 on TID2013, indicating that our random splits are representative of the data and are not outliers. Table 3 shows that our proposed DCNN architecture outperforms the state-of-the-art in both CSIQ and TID2013, except for the PLCC on TID2013 where we are only 0.11% worse than Kim et al. The fact that we are better than (or comparable to) existing methods on the standard datasets (which are smaller and have fewer distortions) while significantly outperforming them in our test set (which contains new distortions) validates our method.

---
[10]Comparisons on other datasets can be found in the supplementary file.
[11]Here, we train our error-estimation function directly on MOS labels as existing datasets do not provide probabilistic pairwise labels. However, this does not change our architecture of $f$ or its number of parameters.

### 5.4 Improving other learning-based IQA methods

As discussed earlier, our pairwise-learning framework is a better way to learn IQA because it has less noise than either subjective human quality scores or Swiss tournaments. In fact, we can use it to improve the performance of existing learning-based algorithms. We observe that since typical FR-IQA methods use a distorted image $A$ and a reference $R$ to compute a quality score, they are effectively an alternative to our error-estimation block $f$ in Fig. 3. Hence, we can replace our implementation of $f(A, R; \theta)$ and $f(B, R; \theta)$ with a previous learning-based IQA method, and then use our pairwise-learning framework to train it with our probability labels. To do this, we duplicate the block to predict the scores for inputs $A$ and $B$, and then subtract the predicted scores and pass them through a sigmoid (i.e., block $h(s_A, s_B)$ in Fig. 3). This estimated probability is then compared to the ground-truth preference for backpropagation.[12]

This enables us to train the *same learning architectures* previously proposed, but with our probabilistic-preference labels. We show results of experiments to train the methods of Kim et al. [25] and Bosse et al. [7] (both FR and NR) in Table 4. By comparing with the corresponding entries in Table 2, we can see that their performance on our test set has improved considerably after our training. This makes sense because probabilistic preference is a better metric and it also allows us to leverage our large, robust IQA dataset. Still, however, our proposed DCNN architecture for error-estimation block $f$ performs better than the existing methods. Finally, the table also shows the performance of these architectures trained on our pairwise-preference dataset, but tested on CSIQ [26] and TID2013 [42], using their MOS labels as ground-truth. While our modifications have improved existing approaches, PieAPP still performs better.

### 6 Conclusion

We have presented a novel, perceptual image-error metric which surpasses existing metrics by leveraging the fact that pairwise preference is a robust way to create large IQA datasets and using a new pairwise-learning framework to train an error-estimation function. Overall, this approach could open the door for new, improved learning-based IQA methods in the future.

---
[12]All details of the training process can be found in the supplementary.